\documentclass{article}

\PassOptionsToPackage{numbers, compress}{natbib}
\usepackage[final]{nips_2017}

\usepackage[utf8]{inputenc} 
\usepackage[T1]{fontenc}    
\usepackage{hyperref}       
\usepackage{url}            
\usepackage{booktabs}       
\usepackage{amsfonts}       
\usepackage{nicefrac}       
\usepackage{microtype}      
\usepackage{graphicx}
\usepackage{rotating}
\title{No Classification without Representation: \\
  Assessing Geodiversity Issues in Open Data Sets \\
  for the Developing World}

\author{
  Shreya Shankar, Yoni Halpern, Eric Breck, James Atwood, Jimbo Wilson, D. Sculley\\
  \texttt{\{shankarshreya, yhalpern, ebreck, atwoodj, jimbo, dsculley\}@google.com}\\
  Google Brain Team\\
}

\begin{document}

\maketitle

\begin{abstract}
  Modern machine learning systems such as image classifiers rely heavily on
  large scale data sets for training.  Such data sets are costly to create,
  thus in practice a small number of freely available, open source data sets
  are widely used. We suggest that examining the {\em geo-diversity}
  of open data sets is critical before adopting a data set for use
  cases in the developing world. We analyze two large, publicly available image
  data sets to assess geo-diversity and find that these data sets appear
  to exhibit an observable amerocentric and eurocentric representation bias.
  Further, we analyze classifiers trained on these data sets to assess the
  impact of these training distributions and find strong differences in the relative performance on images from
  different locales.  These results emphasize the need to ensure
  geo-representation when constructing data sets for use in the developing
  world.
  \end{abstract}

\section{Introduction: Data and the Developing World}

Creating large data sets from scratch can be costly. As such, it is common
for practitioners to use freely available open source data
sets such as ImageNet \cite{imagenet} and Open Images \cite{openimages}
to train vision models. This is particularly desirable when using ML for the
developing world, where resources for creating new data sets may be limited.
However, if these data sets are not {\em representative} of the locations of
interest, predictive performance of models may suffer.

In this paper, we assess the geo-diversity of large data sets and the
differences that models trained
on them exhibit when classifying images from varying
geographical locations. We find an observable amerocentric and
eurocentric bias shown in both forms of assessement.
This is the case despite these data sets' creators efforts to encourage diversity.
We present these findings not as a criticism
but as a case study in the difficulties in creating a geographically balanced
data set.

\section{Background: ImageNet and Open Images}

In this work, we analyze two popular public data sets: ImageNet \cite{imagenet}
and Open Images \cite{openimages}.
These two data sets are generally considered academic benchmarks and are not
necessarily constructed to cover every possible use case.
However, in the absence of a robust data source for a particular application,
it is quite common to fall back to these standard data sets.

The first version of the ImageNet data set was released in 2009 by
Deng \textit{et al.} \cite{imagenet_cvpr09}. An updated 2012 release \cite{imagenet},
used to train the model in this paper, consisted of approximately 1.2 million image thumbnails
and URLs from 1000 categories. Each image in the data set is associated
with a human-verified single label.
The Open Images data set, released in 2016 by Krasin \textit{et al.} \cite{openimages}, contains about 9 million
URLs to Creative Commons licensed images. There are 6012 image-level human-verified labels, and each
image can be associated with multiple labels.

Pretrained image classification models trained on both ImageNet and Open Images
are publicly available on the Tensorflow \cite{tensorflow} Slim\footnote{\url{https://github.com/tensorflow/models/tree/master/research/slim}} and Open Images Github\footnote{\url{https://github.com/openimages/dataset}} pages, respectively.
For each data set, we use publicly released pretrained models with the
Inception V3~\cite{inception} architecture, which gives competitive performance
across standard benchmarks.

\begin{figure}
  \begin{centering}
    \vskip-0.1in
    \includegraphics[width=2.7in]{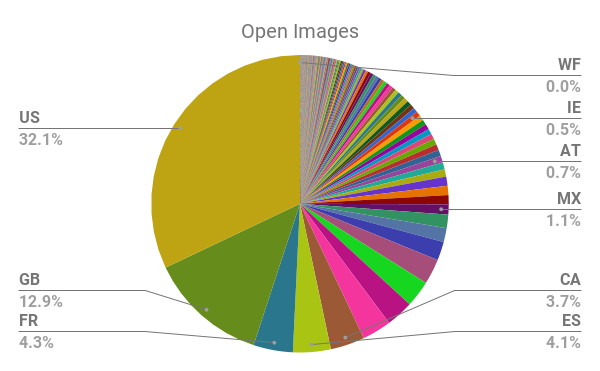}
    \includegraphics[width=2.7in]{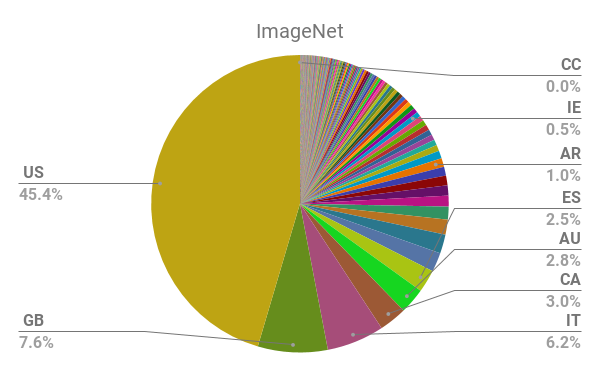}
    \vskip-0.1in
  \caption{\label{images_pie} Fraction of Open Images and ImageNet images
    from each country. In both data sets, top represented locations include the US and Great Britain.
    Countries are represented by their two-letter ISO country codes.}
\end{centering}
\end{figure}

\begin{figure}
  \begin{centering}
    \vskip-0.1in
    \includegraphics[width=4in]{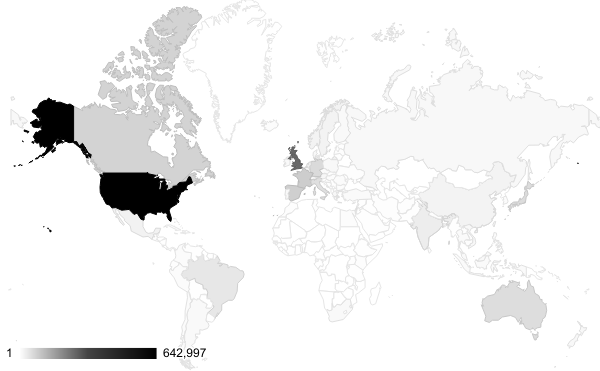}
    \vskip-0.1in
  \caption{\label{open_images_geodiversity} Distribution of the geographically
    identifiable images in the Open Images data set, by country. Almost a third
    of the data in our sample was US-based, and 60\% of the data was from
  the six most represented countries across North America and Europe.}
\end{centering}
\end{figure}

\section{Analyzing Geo-Diversity}
\label{inferred_location}

Our first goal was to assess the geo-diversity of the images in the open source
data sets.
It is naturally difficult to identify the geo-location
of every image in previously released open source image data sets.
However, proxy information
such as textual / contextual information and URL metadata provided by a service
allowed us to recover reasonably reliable
location information at the country level for a
large number of images in each data set.

For the purposes of this study, we take this country
identification, accepting the possibility of noise in the coverage and accuracy of the country-level
geo-location as unlikely to qualitatively impact the larger trends shown.

\paragraph{Geo-Diversity of Open Images.}
Of the 9 million images in the Open Images data set, we were able to
acquire country-level geo-location for roughly 2 million.  This is a large
(but potentially non-uniform) subset of the overall data.  Geo-location
data is shown in Figures ~\ref{images_pie} and ~\ref{open_images_geodiversity}.  Overall,
more than 32\% of the sample data was US-based and 60\% of the data
was from the six most represented countries across North America and Europe.
Meanwhile, China and India -- the two most populous countries in the world --
were represented with only 1\% and 2\% of the images, respectively.
Despite our expectation that there would be some skew, we were surprised
to find this level of imbalance.

\paragraph{Geo-Diversity of ImageNet.}
For the 14 million images in the fall 2011 release of the ImageNet data set,\footnote{\url{http://image-net.org/imagenet_data/urls/imagenet_fall11_urls.tgz}} we
similarly acquired country-level geo-location data. We had lower coverage for ImageNet,
but the distribution was similarly dominated by a small number of countries, as shown in Figure ~\ref{images_pie}.
Around 45\% of the data in our sample was US-based.
Here, China and India were represented with 1\% and 2.1\% of the images, respectively.

\begin{figure}
  \centering
  \vskip-0.2in
  \includegraphics[width=1.85in]{imagenet_grooms_histogram.pdf}
  \includegraphics[width=1.75in]{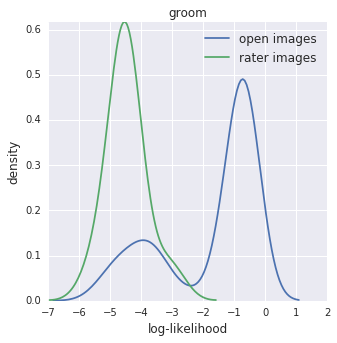}
  \includegraphics[width=1.75in]{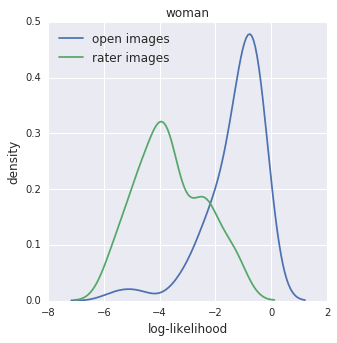}
  \vskip-0.1in
  \caption{Density plots of log-likelihood attributed
    for {\tt groom, bridegroom} images crowdsourced by raters in
    Hyderabad, India, as scored by a model trained on ImageNet (left)
    and Open Images (center), as compared to images in the standard
    test sets.  In both cases, the images provided by Hyderabad-located crowdsourcing are
    dramatically less likely to be recognized correctly by these models.
    The plot on right shows a similar trend for the {\tt woman} class in
    OpenImages which has no corresponding class in ImageNet.
    \label{histograms}
}
\end{figure}

\section{Analyzing Classification Behavior Based on Geo-Location}
We examined how lack of geo-diversity in training data
impacts classification performance on images drawn from a broader
set of locations. We collected image data for specific
geographical regions using two separate methods.

\paragraph{Crowdsourced Data.}
Our first method of collecting stress-test data was to ask crowdsourced
raters located in Hyderabad to find and return URLs of images on the internet
that matched particular labels, specifically
from a community that
they identified with in an effort to avoid amerocentric or eurocentric bias.

Spot checking the results of this
collection showed that images for labels such as {\tt groom, bridegroom}
did appear to represent traditions commonly associated
with India. In other cases the human
raters found it difficult to find an image from a community that they
belonged to. Some of these cases were for US-centric classes,
(e.g., an ``infielder'' baseball player or ``Captain America'').

\paragraph{Geo-located web images.} While the raters in Hyderabad gave us one
source of location-specific image data,
we needed another approach to find data from a wider range of countries.
To this end, we first identified 15 countries to target and joined the
per-country location proxy described above with inferred labels from a
classifier similar to Google Cloud Vision API, across a large data store of images from the web.
For analysis, we focused on labels related to ``people'',
such as bridegroom, police officer, and greengrocer.

One limitation of this work is that even our geographically diverse
images were collected from the internet using tools that rely (at
least partially) on image classifiers themselves. The human raters
used web search to find images that depicted people from their
communities. Similarly, when building a data set from underrepresented countries
using geo-located web images to stress-test a classifier, an image classifier
was used to filter for relevant images.

\paragraph{Geo-Dependent Mis-Classifications.}
Looking over {\tt groom, bridegroom} images supplied by the Hyderabad raters,
we found that the classifier trained on ImageNet data was likely to misclassify
these images as {\tt chain mail}, a kind of armor.
Other images were misclassified as focusing on {\tt cloth},
{\tt academic gown}, or {\tt vestment}.  Using a method similar to
SmoothGrad \cite{smoothgrad}, we looked at saliency maps to determine which parts of the
images were most depended on by the model when making these classifications.
Surprisingly, in all cases that we looked at, the human face in the image was
highlighted rather than the attire, despite the fact that the majority
of misclassifications assigned an attire-based label.

\paragraph{Classifier performance on localized data.}

We use two pretrained models, one trained on ImageNet and another
trained on Open Images to test the difference in classifiers' performances
between data drawn from the standard evaluation data split in ImageNet and
Open Images and rater-supplied images.

\begin{figure}
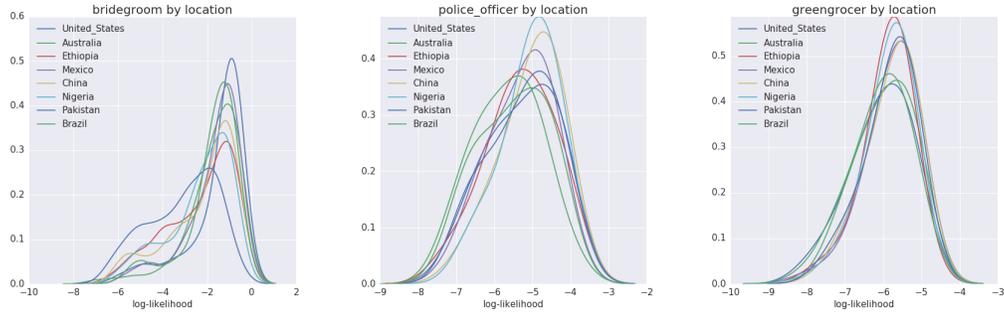

\centering
\vskip-0.1in
\includegraphics[width=1.8in]{bridegroom-histogram.pdf}
\includegraphics[width=1.8in]{police_officer-histogram.pdf}
\includegraphics[width=1.8in]{greengrocer-histogram.pdf}
\vskip-0.1in
\caption{\label{locations} Density plots of log-likelihood attributed by the
models trained on Open Images for images drawn
from the {\tt groom, bridegroom}, {\tt butcher}, {\tt greengrocer}, and {\tt police officer} categories.
Groom images with non-US location tags tend to have lower likelihoods than
the groom images from the US.
\vspace*{-1em}}
\end{figure}

Figure~\ref{histograms} shows some categories that
showed noticeable differences in performance. These differences appear in both
classifiers, suggesting that this problem is not particular to a single data set.
Using the geolocated images from the web, we compare performance
between countries (Figure~\ref{locations}).
Some classes of images have similar distributions of predictions across
countries, indicating that the training data set is better-represented
in such classes.

\begin{figure}
\begin{centering}
\includegraphics[scale=0.6]{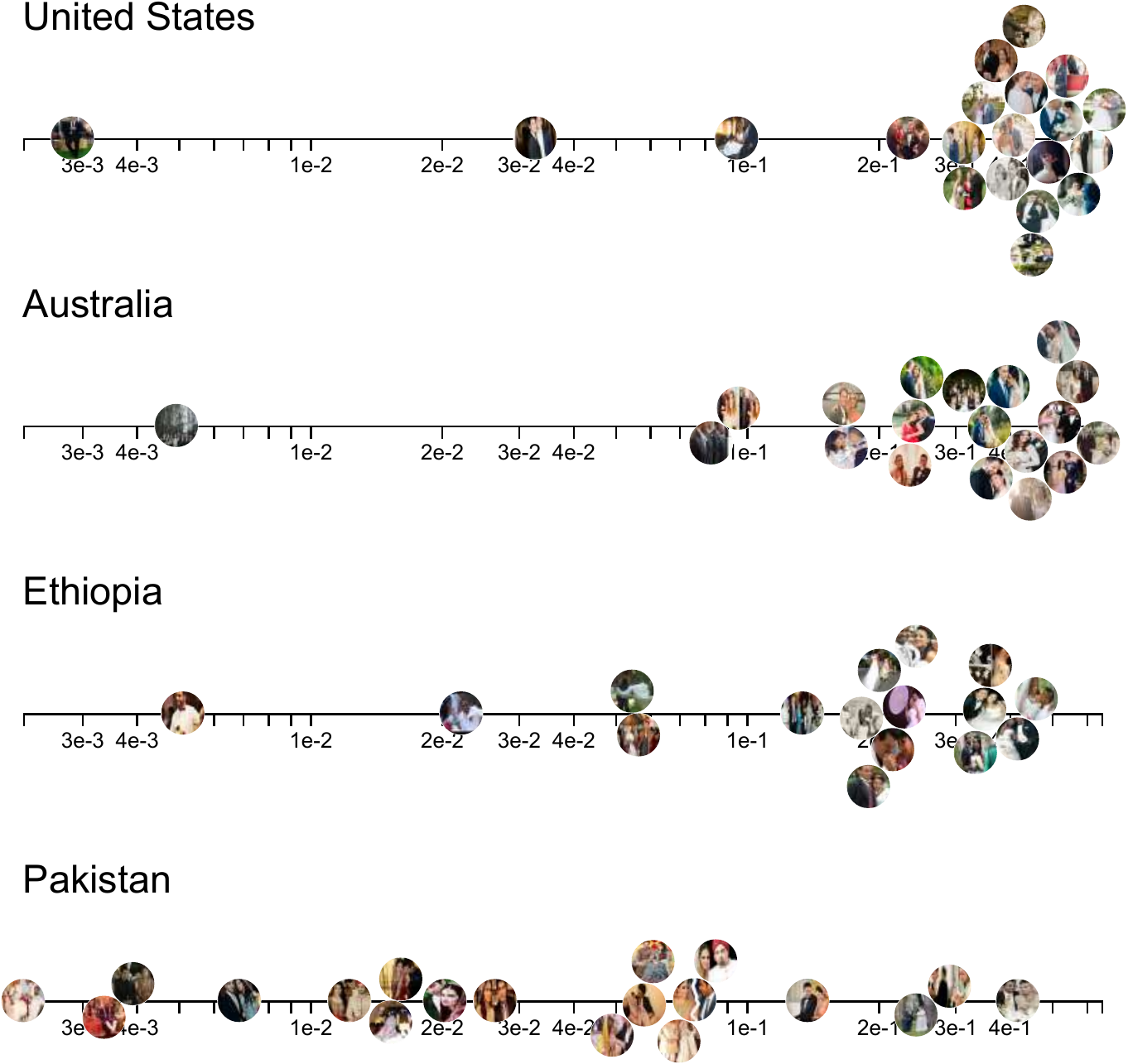}
\caption{\label{groom_photos}
Photos of bridegrooms from different countries aligned by the
log-likelihood that the classifier trained on Open Images assigns to the
bridegroom class.
Images from Ethiopia and Pakistan are not classified as consistently
as images from the United States and Australia.
\vspace*{-1em}}
\end{centering}
\end{figure}

Figure~\ref{groom_photos} plots images of {\tt groom, bridegroom}
images from different countries by log likelihood.  The US-based images
are clustered to the far right, showing high confidence, while images
from Ethiopia and Pakistan are much more uniformly distributed, showing
poorer classifier performance.  We confirmed this trend across several
other countries in different regions of the world.

We focused on labels relating to humans in
this work, but noticeable distributional differences between developed
and developing countries can occur other areas as well, including
sports, transportation, and wildlife.

\section{Discussion}

It is clear that
standard open source data sets such as ImageNet or Open Images may not
have sufficient geo-diversity for broad representation across the
developing world. This is not too surprising, as these data sets
were designed for specific purposes, and it is only the practice of
later adoption for other purposes that may introduce problems.

This study highlights the importance of assessing the appropriateness of a given
data set before using it to learn models for use in the developing world.
Equally, this work emphasizes the importance of creating new data sets that
prioritize broad geo-representation as first class goals,
in order to aid ML in the developing world.

\bibliography{sources}

\end{document}